\documentclass[runningheads]{llncs}
\usepackage[T1]{fontenc}
\usepackage{graphicx,verbatim}

\usepackage{color}

\usepackage{pifont}
\newcommand{\cmark}{\ding{51}}
\newcommand{\xmark}{\ding{55}}
\usepackage{tikz}
\usetikzlibrary{positioning, arrows.meta, fit}
\usepackage{subcaption}
\usepackage{pgfplots}
\pgfplotsset{compat=1.18}
\usepackage{array} 
\newcolumntype{C}[1]{>{\centering\arraybackslash}p{#1}}
\usepackage{multirow}
\usepackage{amsmath}

\begin{document}
\title{CT-CLIP Representations for Multimodal\\ Lung Cancer Survival Prediction}

\author{Sofie Allgöwer\inst{1}\and
Mikael Johansson\inst{2} \and
Andreas Hallqvist\inst{3}\and
Jonas Andersson\inst{2}\and
Åse Johnsson\inst{4}\and
Ida Häggström \inst{1} \and
Jennifer Alvén \inst{1}
}
\authorrunning{F. Author et al.}
%
\institute{Chalmers University of Technology, Gothenburg, Sweden \\ 
\email{allgower@chalmers.se} \and
Umeå University, Umeå, Sweden
\and
Sahlgrenska University Hospital, Gothenburg, Sweden
\and
Sahlgrenska Academy at Gothenburg University, Gothenburg, Sweden
}
    
\maketitle
\begin{abstract}
Accurate prognosis prediction is important for treatment planning in lung cancer, but deep learning-driven survival modelling is often limited by the scarcity of curated imaging cohorts with reliable outcome data.
This study evaluates whether representations from a domain-specific foundation model can be used for multimodal survival prediction in data-constrained clinical settings. 
We assess the foundation model CT-CLIP as a feature extractor for pretreatment computed tomography images and clinical variables from 242 diagnosed lung cancer patients. The evaluation includes adaptation strategies based on frozen encoders, full fine-tuning, and low-rank adaptation, together with modality ablations and comparisons with clinical and multimodal baselines.
The results show that a frozen CT-CLIP model combined with a trainable lightweight survival head outperforms the clinical baseline and achieves comparable or improved performance relative to other multimodal approaches, and separates patients into clinically meaningful high- and low-risk groups.


\end{abstract}
\section{Introduction}
Lung cancer accounts for the highest number of cancer-related deaths worldwide. Despite advances in imaging and treatment, the 5-year survival rate remains below 20\% in most parts of the world~\cite{bray2024global}. This persistently poor prognosis underscores the need for improved risk stratification and prognostication. To personalize clinical decision-making and treatment planning, accurate survival prediction models are of importance. 

One of the primary challenges in developing such models is the limited availability of sufficiently large and well-curated public datasets with imaging, clinical variables, and reliable outcome data. In this work, we use a real-world dataset of 242 diagnosed lung cancer patients, consisting of pretreatment computed tomography (CT) scans and clinical variables, and investigate alternatives for lung cancer survival prediction in data-scarce settings.

\subsubsection{Related work.}
While traditional work performed unimodal survival prediction using methods such as Cox Proportional Hazards (CoxPH)~\cite{coxph} for tabular data, or DeepConvSurv~\cite{zhu2016deepconvsurv} for images, more recent studies have explored multimodal survival prediction. Intuitively, such models may better reflect the prognostication process of medical doctors, who integrate information from multiple modalities when assessing patient outcomes. Several approaches employ Convolutional Neural Networks (CNNs) for CT image feature extraction combined with tabular data through various fusion strategies~\cite{mukherjee2020lungnet,wu2021deepmmsa,stage_I-III_Zheng2023}. However, these models typically rely on manually annotated tumor regions as input.

More recently, Kim et al.~\cite{llm_guided_multimodal_mil} developed a framework for 5-year overall survival prediction, using multiple instance learning to integrate information from CT, pathology images and clinical data. They employed a CNN as image feature extractor, converted tabular data into clinical notes and encoded the notes with CLIP’s~\cite{radford2021_original_clip} pretrained text encoder, which improved the 5-year AUC compared to standard tabular integration. Given the domain gap between medical and general language~\cite{cxr_bert_boecking2022}, it remains unclear whether a text encoder pretrained on medical data would better capture the semantic meaning of clinical notes.

Xing et al.~\cite{cross-modality_xing} presented a lung cancer survival prediction framework with modality-specific encoders: a 3D vision transformer (ViT) for CT volumes and a graph-based transformer for tabular data, followed by a Multilayer Perceptron (MLP). Their work highlights the potential of frozen pretrained encoders to extract meaningful representations from CT volumes and clinical data, but without public code or pretrained weights available at the time of writing, their framework cannot be directly evaluated in our small, real-world clinical cohort.

\subsubsection{Contribution.}
In this work, we present a principled adaptation of the pretrained vision-language foundation model CT-CLIP to multimodal lung cancer survival prediction in a real-world, data-constrained setting. First, we assess whether frozen CT-CLIP representations can support survival modelling from pretreatment CT images and clinical variables. Second, we compare adaptation strategies, including frozen encoders, full fine-tuning, and Low-Rank Adaptation (LoRA). Third, we perform modality ablations and compare against clinical and multimodal baselines, reporting discrimination metrics and risk-group separation. Our results show that frozen foundation-model representations can be competitive with task-specific models for lung cancer survival modelling.

\section{Methodology}

\subsection{Survival analysis}
The objective in survival analysis is to estimate the time until an event of interest occurs, i.e., time-to-death in the context of overall survival.
Based on patient covariates $x$, observed survival time $t$, and a binary indicator $\delta$ distinguishing event occurrence ($\delta=1$) from censoring ($\delta=0$), the relative risk of death can be modeled. In the CoxPH model~\cite{coxph}, the instantaneous risk of death at time $t$ is described by the hazard function,
\begin{equation} \label{eq:risk}
\lambda(t \mid x) = \lambda_0(t) \exp\big(h_\theta(x)\big),
\end{equation}
where $\lambda_0(t)$ is the baseline hazard dependent only on $t$, and $\exp\big(h_\theta(x)\big)$ represents a risk score associated with the patient covariates $x$. 
In CoxPH, the log-risk function $h_\theta(x)$ is estimated linearly as $h_\theta(x)=\theta^Tx$.
Implementations such as DeepSurv~\cite{katzman2018deepsurv} and DeepConvSurv~\cite{zhu2016deepconvsurv}, where neural networks are used to model $h_\theta(x)$, introduce increasing flexibility in the log-risk estimation. 
Cox-like models are optimized by minimizing the average negative partial log-likelihood,

\begin{equation} \label{equation:loss}
\mathcal{L} = - \frac{1}{N_{\delta=1}} 
\sum_{i : \delta_i = 1}
\left(
h_\theta(x_i)
-
\log \sum_{j : t_j \ge t_i}
\exp\big( h_\theta(x_j) \big)
\right),
\end{equation}
where $N_{\delta=1}$ is the number of observed events. While there exist other methods to estimate survival, e.g., DeepHit \cite{lee2018deephit}, we utilize a DeepSurv survival head in this work, as it is a common baseline approach in survival prediction.

\subsection{The vision-language pretraining framework CT-CLIP}
The foundation model CT-CLIP~\cite{hamamci2024ctchat} is an adaptation of the Contrastive Language-Image Pretraining (CLIP) framework~\cite{radford2021_original_clip} to paired chest CT volumes and radiological reports. The model consists of an image encoder and a text encoder, each producing a latent representation of its respective input. More specifically, the vision branch uses the encoder from CT-ViT~\cite{hamamci2024generatect} followed by a linear projection layer, whereas the text branch employs the chest X-ray language model CXR-BERT~\cite{cxr_bert_boecking2022} together with another linear projection layer.

CT-CLIP was trained on CT-RATE \cite{hamamci2024ctchat}, a dataset comprising matched chest CT volumes and radiology reports from 21,304 unique patients. Training is performed using contrastive learning, maximizing similarity between corresponding volume-text pairs while minimizing similarity between non-matching pairs.

Reported zero-shot and fine-tuning experiments demonstrate that CT-CLIP's vision transformer encodes the CT volumes in a clinically meaningful way, and that the learned representations transfer to downstream tasks. Although CT-CLIP was not trained for lung cancer survival prediction, these results suggest that it may provide informative feature representations for survival modelling.

\subsection{Model architecture}
In this work, we adapt CT-CLIP to operate on CT volumes and tabular clinical data, since no radiological reports are available in our dataset. The tabular data is preprocessed into template-based clinical notes to enable the use of CXR-BERT as the tabular data encoder, similar to previous work~\cite{llm_guided_multimodal_mil}.
Although the structure of the generated clinical notes might differ from the radiological reports used to pretrain CT-CLIP, the substantial domain overlap should allow the CXR-BERT encoder to process the information within a similar context.

\begin{figure}[h]
    \centering
    \usetikzlibrary {shapes.geometric}
    \tikzset{
    enc node/.style={trapezium, fill=#1!20, draw=#1!75, text=black}, 
    roundnode/.style={circle, draw=violet!40, fill=violet!10, minimum size=2mm}
    }
    \input{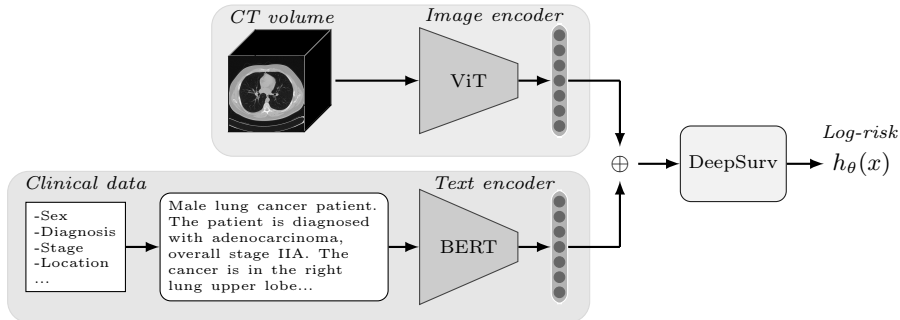}
    \caption{\textit{The CT-CLIP survival framework.} $\oplus$ represents concatenation.}
    \label{fig:nn_overview}
\end{figure}

An overview of the model architecture is presented in Figure~\ref{fig:nn_overview}. The pretrained vision and text transformers from CT-CLIP, including their projection layers, are initialized with pretrained weights and used to create 512-dimensional latent representations of the CT volume and the clinical data, respectively. The resulting embeddings are concatenated into a 1024-dimensional feature vector. This fused representation is passed to a DeepSurv head consisting of two fully connected layers of 512 and 256 units. The final output is a single node estimating the log-risk of death, $h_\theta(x)$, in the CoxPH model, see Eq.~\eqref{eq:risk}. The model is trained using Cox partial log-likelihood, see Eq. \eqref{equation:loss}. Different adaptation strategies,  including frozen encoders, full fine-tuning, and LoRA, are evaluated and described in the experimental section.

\section{Experiments and Results}

\subsection{Dataset}
We curate a dataset comprising pretreatment CT scans and associated clinical variables from 242 patients diagnosed with lung cancer during the years 2008-2018, collected at the University Hospital of Umeå, Sweden. The study was approved by the Swedish Ethical Review Authority (Dnr: 2021-06549-01). Outcome data are collected from Swedish registries. The median survival time in the dataset is 3.4 years. Out of the 242 patients, 189 suffered an event within the time of study. Overall survival is defined as the time from diagnosis to death or last follow-up.

Out of the 242 patients, 124 are female (51.2\%) and 118 male (48.8\%). The diagnoses include adenocarcinoma (53.7\%), squamous cell carcinoma (26.4\%), small cell lung cancer (7.0\%), carcinoid (4.5\%), large cell lung cancer (4.1\%) and other (4.3\%). Tumor stage distribution comprises Stage I (42.9\%), Stage II (13.2\%), and Stage III (43.8\%). In total, we consider the following clinical variables: sex, overall lung cancer stage, T-stage, N-stage, smoking status, diagnosis, ECOG performance status, and location of primary tumor (side and lobe). 

The dataset is split into a held-out test set of 50 patients (20.7\%) and a training set of 192 patients (79.3\%) using stratified sampling, to ensure a similar proportion of censored patients in both groups. The training set is further randomly divided into three folds for cross-validation during model development. The held-out test set is used exclusively for final performance evaluation.

\subsection{Implementation Details}
\subsubsection{Preprocessing.} 
For the CT volumes, the preprocessing pipeline used for pretraining CT-CLIP~\cite{hamamci2024ctchat} is utilized. The 3D volumes are resampled to voxel size $0.75\times0.75\times1.5$ mm and center-cropped or padded to matrix size $480\times480\times240$ voxels. The pixel intensities are converted to Hounsfield units (HU), thresholded to the interval $[-1000,1000]$ HU, and normalized to the range $[-1,1]$. To prepare the clinical data for input to the CT-CLIP text encoder, clinical notes are generated using a manually created template, where corresponding tabular values are imputed, see Figure~\ref{fig:nn_overview}.
To handle missing values, the sentence segment corresponding to that variable is omitted from the notes.

\subsubsection{Adaptation strategies.} 
To explore which training strategy is most beneficial, adaptation experiments are performed. In these, the ViT, BERT, vision projection layer (VP), text projection layer (TP), and DeepSurv head (DS) are treated as individual components, either frozen or fine-tuned. The explored configurations are presented in Table~\ref{tab:freezing_strategies}. Full fine-tuning, where all model weights are updated, and parameter-efficient fine-tuning using LoRA are performed for all configurations. LoRA is implemented using the PEFT library~\cite{peft} and configured with rank $r=8$, based on optimal rank experiments in~\cite{hu2022lora}, and $\alpha=2\times r=16$, the typical choice in~\cite{peft}. LoRA adapters are applied to the query and value projection matrices in BERT, to the query, key and value matrices in ViT and to the linear projection layers VP and TP. DS is always trained from scratch.  

\subsubsection{Training.} 
The models are trained for a maximum of 100 epochs with early stopping applied after 10 consecutive epochs without improvement in the validation loss. Optimization is performed using Adam with $\beta_1=0.9$, $\beta_2=0.999$, and weight decay $1\times10^{-5}$. The learning rate, selected from $\in\{1\times10^{-4}, 5\times10^{-5}, 1\times10^{-5}, 5\times10^{-6}, 1\times10^{-6}\}$, is determined via grid search on the validation folds, selecting the value that yields the highest mean performance across the validation folds. For each fold, the model weights corresponding to the epoch with the highest C-index (Harrell's) on the validation set are selected. Pretrained weights of version \verb!CT-CLIP_v2.pt! are utilized. The DS head employs ReLU activations and a dropout rate of 0.25. The framework was implemented in PyTorch, and one NVIDIA RTX A6000 GPU was used for the experiments.

\subsubsection{Baseline models.}
The reference multimodal models are trained using the official implementations provided in~\cite{daft_Wolf2022}.
As Interactive-Model and FiLM are not originally developed for survival prediction, adapted versions are utilized. ResNet+Tabular, corresponding to the Concat-2FC implementation in~\cite{daft_Wolf2022}, represents a common approach for multimodal survival prediction, combining a CNN feature extractor with tabular clinical data and an MLP survival head. For all models, missing variables are imputed with the most frequent class.

\subsection{Results}

\subsubsection{Adaptation experiments.}
Results on the validation set from the adaptation experiments, evaluated using Harrell's C-index, are presented in Table~\ref{tab:freezing_strategies}. 
The results indicate that configurations with frozen projection layers may have better discriminative ability than their trainable counterparts, suggesting that the pretrained projection layers successfully map the image and text embeddings into a shared latent space, potentially informative for survival prediction.
However, the results on the validation set do not provide clear evidence that either full fine-tuning or LoRA is preferable for adapting CT-CLIP to the survival prediction task.

\begin{table}[tb]
\centering
\caption{\textit{Adaptation experiments.} 3-fold cross-validation results (mean$\pm$std) on the validation folds. Trainable components are marked with \cmark. The best-performing models are shown in bold.}\label{tab:freezing_strategies}
\fontsize{9}{12}\selectfont
\begin{tabular}{C{1cm}C{1cm}C{1cm}C{1cm}C{1cm}|C{2.5cm}|C{2.5cm}}
\hline
\multicolumn{5}{c|}{Trainable components} & \multicolumn{2}{c}{Harrell's C} \\ \hline
ViT & BERT & VP & TP & DS  & Fine-tuned & LoRA \\ \hline 
\multicolumn{1}{c}{\cmark} & \cmark & \cmark & \cmark & \multicolumn{1}{c|}{\cmark} & $0.675\pm0.035$ & $0.701\pm0.038$ \\
\multicolumn{1}{c}{\cmark} & \cmark &  &  & \multicolumn{1}{c|}{\cmark} & $\mathbf{0.715\pm0.019}$ & $\mathbf{0.715\pm0.020}$ \\ \hline
\multicolumn{1}{c}{\cmark} &  & \cmark & \cmark & \multicolumn{1}{c|}{\cmark} & $0.703\pm0.035$ & $0.698\pm0.032$ \\
\multicolumn{1}{c}{\cmark} &  & \cmark &  & \multicolumn{1}{c|}{\cmark} & $0.700\pm0.039$ & $0.697\pm0.046$ \\
\multicolumn{1}{c}{\cmark} &  &  &  & \multicolumn{1}{c|}{\cmark} & $0.714\pm0.033$ & $0.708\pm0.032$ \\ \hline
\multicolumn{1}{c}{} & \cmark & \cmark & \cmark & \multicolumn{1}{c|}{\cmark} & $0.686\pm0.026$ & $0.712\pm0.039$ \\
\multicolumn{1}{c}{} & \cmark &  & \cmark & \multicolumn{1}{c|}{\cmark} & $0.705\pm0.034$ & $0.710\pm0.033$ \\
\multicolumn{1}{c}{} & \cmark &  &  & \multicolumn{1}{c|}{\cmark} & $0.704\pm0.032$ & $0.713\pm0.035$ \\ \hline
\multicolumn{1}{c}{} &  & \cmark & \cmark & \multicolumn{1}{c|}{\cmark} & $0.707\pm0.034$ & $0.701\pm0.046$ \\  
\multicolumn{1}{c}{} &  &  &  & \multicolumn{1}{c|}{\cmark} & $0.711\pm0.030$ & $0.711\pm0.030$  \\ \hline
\end{tabular}
\end{table}

\subsubsection{Modality ablation.}
Ablation experiments using an image-only or text-only branch together with the DS head are performed to explore the contribution of each modality branch in CT-CLIP. The unimodal experiments follow the same fine-tuning scheme as presented in Table~\ref{tab:freezing_strategies}. For each model type, three variants are evaluated on the held-out test set, the one with a frozen backbone and the best-performing models on the validation set with full fine-tuning and LoRA. 

As shown in Table~\ref{tab:clip_model_ablation}, the multimodal model achieves better performance than its unimodal counterparts.
Moreover, a consistent pattern across model types indicates that freezing the backbone leads to improved discriminative ability compared to full fine-tuning, and to comparable or improved performance to LoRA. 
This may indicate overfitting when adapting the models to limited training data, and highlights the benefit of utilizing the CT-CLIP backbone in a frozen setting for lung cancer survival prediction. 

\begin{table}[tb]
\centering
\caption{\textit{Modality ablation.} Evaluation on the held-out test set ($N=50$). Results reported as mean$\pm$std across the three cross-validation folds. The best-performing model is shown in bold. No trainable model performed significantly better than its frozen counterpart, according to Bonferroni-corrected t-tests at $\alpha = 0.05$.}\label{tab:clip_model_ablation}
{
\fontsize{9}{12.5}\selectfont
\begin{tabular}{p{3.4cm}C{1.2cm}C{1.2cm}C{2.4cm}C{2.4cm}}
\hline
Model type & Image & Tabular & Backbone & Harrell's C \\ \hline
\multirow{3}{10em}{\textit{CT-CLIP+DeepSurv}} & \multicolumn{1}{|c}{\cmark} & \multicolumn{1}{c|}{\cmark} & frozen & \multicolumn{1}{|c}{$\mathbf{0.755\pm0.006}$} \\
 & \multicolumn{1}{|c}{\cmark} & \multicolumn{1}{c|}{\cmark} & fine-tuned & \multicolumn{1}{|c}{$0.692\pm0.025$} \\ 
 & \multicolumn{1}{|c}{\cmark} & \multicolumn{1}{c|}{\cmark} & LoRA & \multicolumn{1}{|c}{$0.746\pm0.006$} \\ \hline
\multirow{3}{10em}{\textit{ViT+VP+DeepSurv}} & \multicolumn{1}{|c}{\cmark} & \multicolumn{1}{c|}{\xmark} & frozen & \multicolumn{1}{|c}{$0.741\pm0.019$} \\ 
 & \multicolumn{1}{|c}{\cmark} & \multicolumn{1}{c|}{\xmark} & fine-tuned & \multicolumn{1}{|c}{$0.679\pm0.015$} \\ 
  & \multicolumn{1}{|c}{\cmark} & \multicolumn{1}{c|}{\xmark} & LoRA & \multicolumn{1}{|c}{$0.744\pm0.005$} \\ \hline 
\multirow{3}{10em}{\textit{BERT+TP+DeepSurv}} & \multicolumn{1}{|c}{\xmark} & \multicolumn{1}{c|}{\cmark} & frozen & \multicolumn{1}{|c}{$0.713\pm0.002$} \\
 & \multicolumn{1}{|c}{\xmark} & \multicolumn{1}{c|}{\cmark} & fine-tuned & \multicolumn{1}{|c}{$0.676\pm0.016$} \\ 
 & \multicolumn{1}{|c}{\xmark} & \multicolumn{1}{c|}{\cmark} & LoRA & \multicolumn{1}{|c}{$0.715\pm0.011$} \\ 
\hline
\end{tabular}
}
\end{table}

\subsubsection{Comparison with baseline and multimodal approaches.}
Table \ref{tab:comparision_models} shows the best-performing CT-CLIP survival model compared to the clinical baseline CoxPH and four other multimodal survival prediction methods. 
The frozen CT-CLIP-based model outperforms CoxPH and the reference multimodal models reported by Xing et al. \cite{cross-modality_xing} across all evaluated metrics, while achieving comparable performance to ResNet+Tabular. However, note that the CT-CLIP-based model is statistically separated from the tabular baseline, in contrast to ResNet+Tabular. No direct comparison with \cite{cross-modality_xing} is included due to the lack of publicly available code.

\begin{table}[h] 
\centering
\caption{\textit{Comparison with baseline and multimodal approaches.} Results on the held-out test set ($N=50$), reported as mean$\pm$std across the three cross-validation folds. $(^*)$ indicates significant difference compared to the tabular baseline (CoxPH) using a Bonferroni-corrected t-test at $\alpha = 0.05$. 
Best results in bold font and second to best underlined.}\label{tab:comparision_models}
{
\fontsize{9}{14}\selectfont 
\begin{tabular}{p{3.0cm}C{0.5cm}C{0.7cm}C{2.4cm}C{2.4cm}C{2.1cm}}
\hline
Model & Img & Tab & Harrell's C & Uno's C & 5-year AUC \\ \hline
\multicolumn{1}{l|}{CT-CLIP (\textit{ours})} & \cmark & \multicolumn{1}{c|}{\cmark} & \underline{$0.755\pm0.006^*$} & \underline{$0.744\pm0.006^*$} & $\mathbf{0.853\pm0.016}$ \\
\multicolumn{1}{l|}{Interactive-Model \cite{Interactive_model_Duanmu2020}} & \cmark & \multicolumn{1}{c|}{\cmark} & $0.741\pm0.007$  & $0.732\pm0.010$ & $0.829\pm0.028$ \\
\multicolumn{1}{l|}{DAFT \cite{daft_Wolf2022}} & \cmark & \multicolumn{1}{c|}{\cmark} & $0.725\pm0.032$  & $0.716\pm0.033$ & $0.837\pm0.026$ \\
\multicolumn{1}{l|}{FiLM \cite{film_perez2018}} & \cmark & \multicolumn{1}{c|}{\cmark} & $0.728\pm0.016$  & $0.720\pm0.016$ & $0.794\pm0.016$ \\
\multicolumn{1}{l|}{ResNet+Tabular} & \cmark & \multicolumn{1}{c|}{\cmark} & $\mathbf{0.757\pm0.015}$  & $\mathbf{0.747\pm0.019}$ & \underline{$0.844\pm0.023$} \\ \hline
\multicolumn{1}{l|}{CoxPH} & \xmark & \multicolumn{1}{c|}{\cmark} & $0.721\pm0.007$ & $0.713\pm0.007$ & $0.772\pm0.032$  \\ \hline
\end{tabular}
}
\end{table}

\subsubsection{Risk stratification.}
\begin{figure}
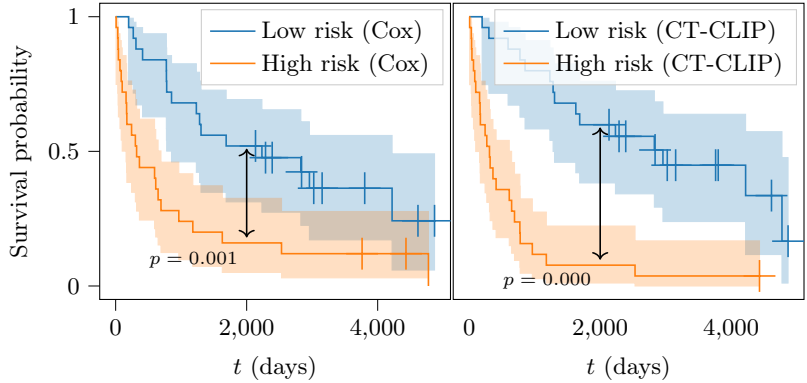

    \centering
    \newsavebox\Cox
\sbox\Cox{\input{cox}}

\newsavebox\CLIP
\sbox\CLIP{\input{clip}}

\begin{tikzpicture}
        \node[inner sep=0pt](E){\usebox{\Cox}};
        \node(T)[right= -0.1cm of E.south east, anchor = south west,inner sep=0pt]{\usebox{\CLIP}};   
\end{tikzpicture}    
    \caption{Kaplan-Meier curves for CoxPH and CT-CLIP from the best fold.}
    \label{fig:kaplanmeier}
\end{figure}

Finally, we demonstrate that the CT-CLIP survival prediction model can be used to divide patients into high- and low-risk groups, by plotting the Kaplan-Meier estimates of the survival function, see Figure~\ref{fig:kaplanmeier}. The median predicted risk is used to stratify patients into high- and low-risk groups and log-rank tests show a significant separation between groups ($p<0.001$).
Compared to CoxPH, the CT-CLIP model shows better separation across all folds.

\section{Conclusion}
We present convincing evidence for leveraging a domain-specific foundation model on a real-world lung cancer survival prediction task. By combining pretrained image and text encoders with a survival head, the CT-CLIP-based model outperforms the clinical baseline and achieves comparable or improved performance relative to other multimodal, task-specific approaches. These findings indicate that pretrained vision-language representations can serve as informative feature extractors in multimodal survival analysis. Importantly, the fully frozen CT-CLIP configuration achieved the best performance in our experiments. Restricting training to a lightweight survival head substantially reduces the number of trainable parameters, which may mitigate overfitting and lower computational cost, thereby facilitating use in real-world, low-resource settings.

\subsubsection{Acknowledgment.} The study was funded by the Sjöberg Foundation grant 2022-489.

%
%
\bibliographystyle{splncs04}
\bibliography{mybibliography}

\end{document}